# Behavioral Conflict Avoidance Between Humans and Quadruped Robots in Shared Environments*

Shuang Wei[1], Muhua Zhang[1], Yun Gan[1], Deqing Huang[1], Lei Ma[1] and Chenguang Yang[2], *Fellow, IEEE*

*Abstract*—Nowadays, robots are increasingly operated in environments shared with humans, where conflicts between human and robot behaviors may compromise safety. This paper presents a proactive behavioral conflict avoidance framework based on the principle of adaptation to trends for quadruped robots that not only ensures the robot's safety but also minimizes interference with human activities. It can proactively avoid potential conflicts with approaching humans or other dynamic objects, whether the robot is stationary or in motion, then swiftly resume its tasks once the conflict subsides. An enhanced approach is proposed to achieve precise human detection and tracking on vibratory robot platform equipped with low-cost hybrid solid-state LiDAR. When potential conflict detected, the robot selects an avoidance point and executes an evasion maneuver before resuming its task. This approach contrasts with conventional methods that remain goal-driven, often resulting in aggressive behaviors, such as forcibly bypassing obstacles and causing conflicts or becoming stuck in deadlock scenarios. The selection of avoidance points is achieved by integrating static and dynamic obstacle to generate a potential field map. The robot then searches for feasible regions within this map and determines the optimal avoidance point using an evaluation function. Experimental results demonstrate that the framework significantly reduces interference with human activities, enhances the safety of both robots and persons.

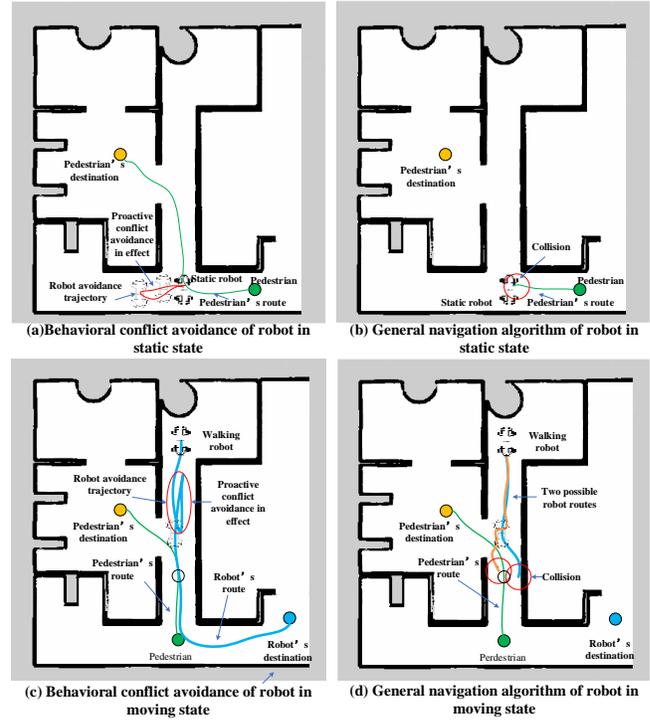

Fig. 1. Comparison chart with and without behavioral conflict avoidance.

## I. INTRODUCTION

With the rapid advancement of robotics technology, mobile robots are increasingly integrated into industrial, service, and domestic settings that involve close human–robot interactions. However, such shared environments often lead to conflicts between robotic and human actions. For example, as shown in **Figure 1(b)**, stationary robots without a clear target often find it difficult to develop effective responses when encountering dynamic obstacles. In these cases, the sudden appearance of a human or another dynamic object forces the robot to remain passive, compelling humans to alter their behavior to avoid a conflict. Moreover, as shown in **Figure 1(d)**, when robots in motion encounter unexpected dynamic obstacles, conventional approaches typically rely on real-time path re-planning based on a globally planned route. This re-planning process is inherently goal-directed but may lead to overly aggressive maneuvers, such as abrupt detours that could provoke conflicts, or even situations where the robot becomes immobilized due to a lack of viable paths, especially in constrained spaces like narrow corridors. Repeated occurrences of such behaviors can diminish human trust in robotic systems and pose significant safety risks for both parties.

To address these challenges while ensuring the robot's safety and minimizing its interference with human actions, prior studies [3], [4] have adopted a detect-avoid-recover mechanism to handle predicted conflicts or imminent conflicts, a concept often referred to as 'trend adaptation.' However, these methods are typically limited to local obstacle avoidance. Building on these approaches, this paper presents an algorithm specifically designed for quadruped robots to prevent human-robot conflicts. The proposed method consists of the following components:

- **Perception and Dynamic Detection:** Utilizing a low-cost MID360 LiDAR, the system enhances the Dynablox algorithm to meet the challenges posed by the high-vibration dynamics of quadruped robots. This enables rapid and accurate detection of dynamic objects. Additionally, an Extended Kalman Filter (EKF) is applied to predict the trajectories of detected

* This research was jointly supported by the Central Government Guides Local Science and Technology Development Project of Sichuan Province under Grant 2024ZYD0018.
[1] School of Electrical Engineering, Southwest Jiaotong University, Chengdu 611756, China ({ws12138@my., muhua.zhang@my., 2022200492@my., elehd@home., malei@wy2022@my.}swjtu.edu.cn).
[2] Department of Computer Science, University of Liverpool, Liverpool, L69 3BX, UK．(cyang@ieee.org).

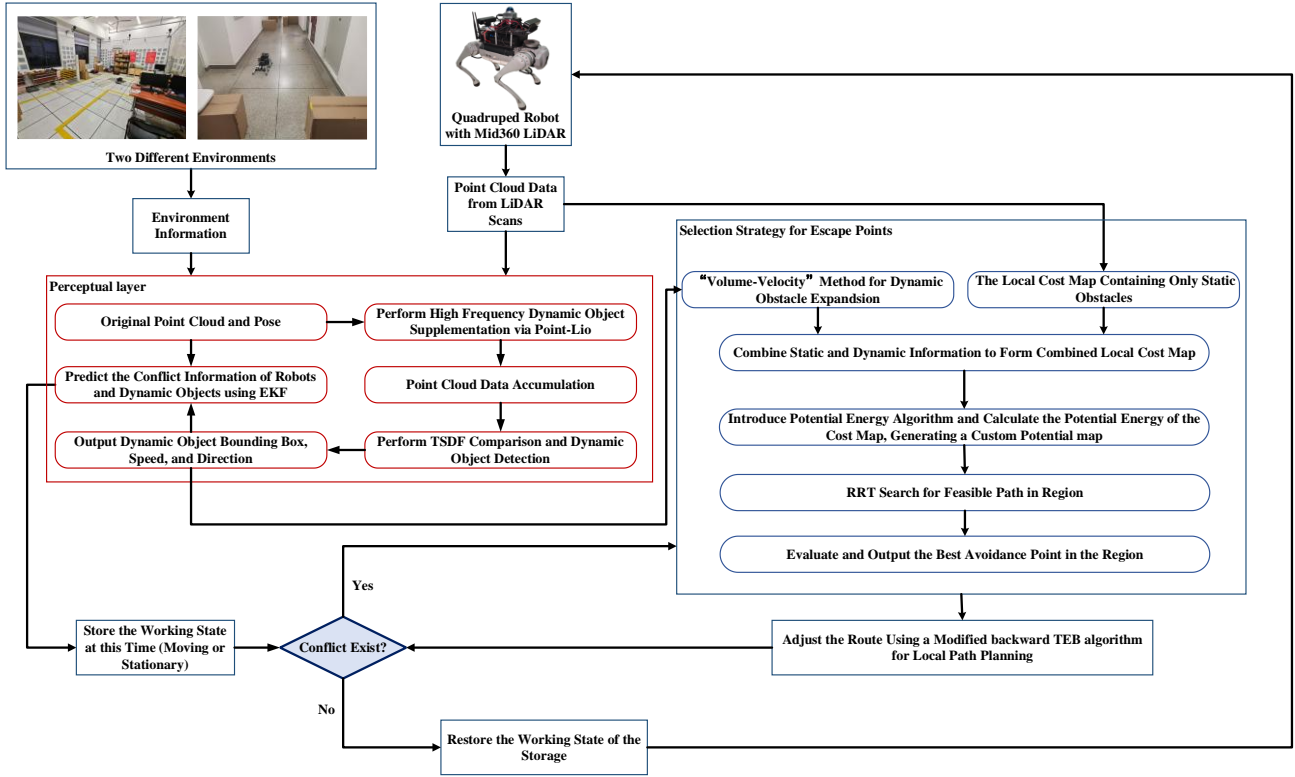

Fig. 2. Frame diagram of behavioral conflict avoidance between humans and quadruped robots.

obstacles, thereby facilitating early assessment of potential conflict risks.

- **Avoidance Strategy:** In scenarios with predicted conflicts, the system constructs a potential field map centered on the robot by fusing information from static obstacles (sensed via LiDAR) and dynamic obstacles (analyzed through a "volumetric speed" approach). Based on this map, a real-time RRT search identifies the current feasible domain, and an optimal avoidance point is determined through a comprehensive evaluation that considers potential field values, path cost, and distance penalties.

- **Local Path Planning and Task Recovery:** An enhanced forward-backward Timed Elastic Band (TEB) algorithm is employed to plan a local path toward the chosen avoidance point, ensuring that the robot's maneuvers are both smooth and contextually appropriate. Once the conflict is resolved, the robot promptly resumes its original task trajectory.

As illustrated in **Figure 1(a)**, when a stationary robot detects an imminent conflict, it autonomously moves in the direction opposite to the approaching pedestrian, and then returns to its original position once the risk is eliminated. Similarly, as shown in **Figure 1(c)**, a robot navigating through a narrow space may temporarily retreat until the hazard subsides, before resuming its planned route. This "go with the flow" strategy not only ensures the robot's safety and effective task execution but also minimizes the disruption to human activities, thereby promoting a harmonious human–robot coexistence.

## II. RELATED WORK

### A. Dynamic Object Detection

In shared human–robot environments, it is crucial that robots detect impending conflicts both promptly and accurately, including the effective detection and tracking of dynamic objects. In [5], [6], image-based methods are employed for dynamic object detection and tracking; however, these approaches are limited to detecting specific objects and rely on fixed viewpoints, thereby reducing their adaptability for mobile robots. Methods based on LiDAR point clouds can address these limitations more effectively. For instance, studies [7]–[9] utilize appearance-based detection techniques, but these require training with pre-calibrated examples or models, assume a flat terrain, and are unable to detect unknown moving objects, making them unsuitable for many robotic applications. Although several learning-based methods have been shown to achieve relatively accurate dynamic object segmentation [10], [11], they operate too slowly to support online real-time applications. In [1], the authors state, " To achieve robust moving object detection in diverse and unstructured environments, we propose an online mapping-based approach that is independent of object appearance, leveraging motion cues extracted from state-of-the-art volumetric mapping solutions." However, the LiDAR employed in this study is relatively expensive, and the approach has not been validated on actual robots, nor has it considered the impact of the robot's own motion on dynamic object detection.

### B. Conflict Avoidance

Most existing algorithms handle dynamic obstacles by prioritizing obstacle avoidance along target-driven paths,

ensuring the robot's safety. Methods such as Velocity Obstacles (VO) [12], [13], Artificial Potential Fields (APF) [14], [15], and the TEB approach [16] exemplify this strategy. However, these methods typically operate frame by frame, considering only the obstacles detected at each moment, making it difficult to consistently and effectively avoid moving obstacles. While Model Predictive Control (MPC) [17], [18] has been widely used for dynamic obstacle avoidance, many approaches struggle when the robot fails to find a feasible path to its target, often leading to unintended interference with moving humans. Additionally, dynamic object motion is often oversimplified, commonly assuming a constant velocity [19] or using basic motion models with fixed velocity and acceleration predictions [20], [21]. In [3], a "retreat to advance" strategy is introduced, allowing the robot to continuously retreat when no viable path exists, offering insights for our approach. Most studies take a robot-centric view, focusing on avoiding collisions but overlooking how the robot's behavior influences human movement. These approaches emphasize task completion and navigation efficiency without considering their impact on human decision-making. This limitation is even more evident when the robot is stationary without a clear navigation goal—an often-overlooked scenario where proactive conflict avoidance is crucial yet largely unaddressed in current research

### III. CONTRIBUTIONS

This research presents a comprehensive framework that addresses potential conflicts encountered by quadruped robots in human-shared environments. By integrating innovative perception, decision-making, and execution mechanisms, the proposed system enables robots to perform tasks efficiently in complex scenarios while proactively avoiding conflicts and minimizing interference with human activities. The main contributions are as follows:

- **Improved Online Dynamic Object Detection Method:** To tackle the degradation in dynamic object detection accuracy under high-vibration conditions, this study systematically enhances existing algorithms. First, Point-LIO [2] technology is introduced at the front end to effectively suppress sensor noise. Then, a backend module integrating bounding box detection and centroid tracking is developed, culminating in a perception system based on the low-cost MID360 LiDAR. This system can rapidly and robustly detect and track dynamic obstacles even in high-vibration environments, providing precise perception inputs for subsequent prediction and conflict avoidance decisions.

- **Innovative Avoidance Point Selection Strategy:** To effectively handle scenarios involving both static obstacles and dynamic conflicts in complex environments, a novel avoidance point selection strategy is proposed. This strategy fuses static obstacle data obtained from LiDAR scans with dynamic obstacle information predicted using the "volumetric speed" method to generate a unified potential field map. Based on this map, the system employs an online Rapidly-exploring Random Tree (RRT) search in conjunction with a custom evaluation function—which considers potential field values, path cost, and distance penalties—to select the optimal avoidance point in real time. Unlike fixed approaches, such as following a predetermined Bézier curve [4], this method enables the robot to yield based on comprehensive environmental information, resulting in a more measured response during human conflict scenarios.

- **Proactive Conflict Avoidance Mechanism:** A complete proactive conflict avoidance mechanism is developed to ensure the robot effectively avoids conflicts with dynamic obstacles whether stationary or in motion. This mechanism promptly activates the avoidance strategy upon detecting potential conflict risks and then allows the robot to swiftly resume its original task—either by returning to its initial position or by continuing toward a predetermined goal—once the risk subsides. In doing so, the system not only ensures the safety of both the robot and nearby humans but also minimizes the robot's disruptive impact on human behavior while enhancing task execution efficiency.

### IV. FRAMEWORK DESIGN

**Figure 2** illustrates the overall framework. Whether operating in a stationary or mobile state within a known-map environment, the robot's perception layer continuously monitors its surroundings for dynamic objects and assesses potential conflict risks. Upon detecting a risk, the system selects an optimal avoidance point by integrating a potential field map—constructed from both static and dynamic obstacles—with an RRT search strategy. Then, an enhanced TEB algorithm executes a context-aware and adaptive avoidance maneuver that enables the robot to safely avoid conflicts and seamlessly resume its original task once the risk subsides.

*A. Dynamic Object Detection*

High-speed robotic motion and sensor jitter often lead to substantial pose estimation errors, which degrade map quality and compromise subsequent dynamic object detection. To address this, we build upon DynaBlox [1]—which identifies moving points by detecting changes in high-confidence free-space within a truncated signed distance field (TSDF)—and incorporate Point-LIO [2] to achieve robust high-frequency odometry and reduced frame-to-frame distortion. By fusing these methods, we maintain the object-agnostic advantages of DynaBlox's free-space approach, while ensuring reliability even when facing large angular velocities and inertial measurement unit (IMU) saturation. Finally, the dynamic point cloud obtained is used to form individual dynamic object bounding boxes through tracking, and the centroids and velocities of these bounding boxes are then outputted. The specific process is shown in **Figure 3**.

DynaBlox operates by incrementally constructing a TSDF from LiDAR or depth sensor data. A point has $x \in R^3$ a signed distance $SDF(x)$ to its nearest surface; this value is then truncated to bound both memory usage and computation:

$$TSDF(x) = \begin{cases} SDF(x), & |SDF(x)| \leq \tau \\ \text{sgn}(SDF(x))\tau, & \text{otherwise} \end{cases}. \quad (1)$$

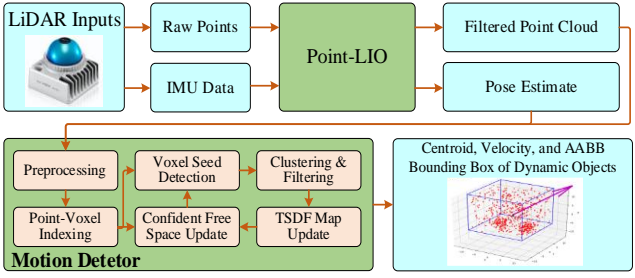

Fig. 3. Flow chart of dynamic object detection.

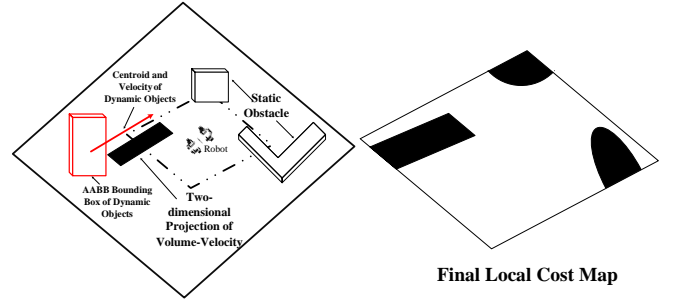

Fig. 4. The dynamic information and local information are integrated to form a new local cost map.

where $\tau$ is a threshold parameter that controls the maximum distance from the surface at which the signed distance is preserved. Over time, TSDF values are updated with newly observed points. Let $\text{TSDF}_t(v)$ and $\omega_t(v)$ be the TSDF value and weight at voxel $v$. Upon receiving a new measurement $\text{TSDF}_{new}$ with $weight_{\omega_{new}}$, the update follows:

$$\text{TSDF}_{t+1}(v) = \frac{\text{TSDF}_t(v)\omega_t(v) + \text{TSDF}_{new}\omega_{new}}{\omega_t(v) + \omega_{new}}$$
$$\omega_{t+1}(v) = \omega_t(v) + \omega_{new} \quad (2)$$

Once this volumetric map is established, DynaBlox robustly labels a voxel $v$ as high-confidence free if it remains consistently unoccupied for enough frames and passes local spatial checks. If a new LiDAR point $p_i$ is found in a voxel previously tagged free, the system concludes that a moving object has arrived, marking $p_i$ as dynamic without requiring any prior appearance model.

To ensure the TSDF remains accurate under rapid robot motion, we adopt Point-LIO. Unlike typical frame-based odometry that integrates the IMU as a control input across an entire scan period, Point-LIO treats the IMU readings—angular velocity $\omega_{pointlio}$ and acceleration $a_{pointlio}$—as measurements of a stochastic process. Each LiDAR point triggers a state update at its actual timestamp, curtailing motion distortion and significantly increasing the update rate. If the IMU range is exceeded, LiDAR residuals still guide the filter, preserving pose stability. By supplying near-real-time pose estimates $T(t)$ Point-LIO prevents the map from drifting excessively, even when the low-cost LiDAR faces sparse returns or high jitter.

After DynaBlox designates which points are dynamic (based on their encroachment into previously free voxels), we aggregate them into clusters to represent distinct moving objects. Let $P_{dyn}^{(t)}$ be the set of dynamic points at time $t$. We define a Euclidean clustering rule, for example:

$$C_\iota = \{p_i | p_i \in P_{dyn}^{(t)}, \exists p_j \in C_\iota : \|p_i - p_j\| \leq \delta\}, \quad (3)$$

where $\|\cdot\|$ denotes the Euclidean distance in 3D and $\delta$ is a spatial threshold. This step consolidates scattered dynamic measurements into object-level candidates $\{C_\iota\}$. When tracking them across frames, a centroid-based distance metric may be used:

$$d(C_\iota^{(t)}, C_m^{(t-1)}) = \|\bar{p}(C_\iota^{(t)}) - \bar{p}(C_m^{(t-1)})\|, \quad (4)$$

where $\bar{p}(C)$ is the centroid of cluster $C$. Those pairs of clusters whose centroids lie within a matching threshold are deemed to represent the same moving object persisting over time. This multi-frame linkage reduces spurious detections while producing short-term motion trajectories, beneficial for conflict avoidance or path forecasting.

By combining the free-space cue in DynaBlox with the high-bandwidth odometry of Point-LIO, we obtain a system robust to fast rotations, sensor jitter, and partial IMU saturation. DynaBlox's ability to detect motion without explicit appearance models, together with point-wise LiDAR updates, preserves consistent map quality and avoids frame-level motion distortion. The system then groups dynamic measurements into clusters, forming preliminary object definitions and enabling multi-frame tracking.

*B. Avoidance Point Selection Strategy*

The next phase of our system involves utilizing an EKF to predict the trajectories of dynamic obstacles. Upon detecting a potential conflict, the robot calculates a safe region, defined as an optimal avoidance point, to temporarily deviate from its planned trajectory.

In constructing the potential map, the robot first incorporates static obstacles derived from low-cost LiDAR scans, such as walls or stationary objects in the surrounding environment. Next, for each newly detected dynamic obstacle, we apply a lightweight EKF to estimate its short-term position $x_{obj}(t + \Delta t)$ and velocity $v_{obj}(t + \Delta t)$. The system gains approximate information on the obstacle's near-future pose. Building on these estimates, we use a "Volume × Velocity" method to dilate the obstacle's bounding shape $B(t)$ in the direction of its predicted motion, as it was showed in **Figure 4**. This procedure marks the corresponding grid cells with high cost, ensuring the planner avoids those dynamically expanding regions. By merging static and dynamic obstacles, we derive a fused local map whose cells are either traversable or identified as high-cost if they fall within—or too close to—an obstacle's envelope.

To obtain a scalar metric that drives the robot away from hazards and attracts it to viable goals, we define a total potential function

$$U(x) = U_{att}(x) + U_{rep}(x). \quad (5)$$

Here, $x \in \mathbb{R}^2$ represents a location in the local map, and $x_g$ denotes a relevant goal position (for instance, the robot's navigation target or a fallback point during conflict alerts). The attractive component encourages

$$U_{att}(x) = \frac{1}{2}\alpha\|x - x_g\|^2 \quad (6)$$

movement toward $x_g$, where $\alpha$ controls the strength of this pull. The repulsive component is given by

$$U_{rep}(x) = \begin{cases} \frac{1}{2}\beta\left(\frac{1}{d(x)} - \frac{1}{d_0}\right)^2, & \text{if } d(x) \le d_0 \\ 0, & \text{otherwise} \end{cases} \quad (7)$$

where $d(x)$ is the distance from $x$ to the nearest obstacle boundary (including the expanded dynamic regions), $d_0$ is a threshold distance beyond which the obstacle exerts no influence, and $\beta$ determines how sharply the potential increases as the robot nears obstacles.

In order to improve the computational speed and allow for faster robot responses, we also incorporate dynamic object information into the goal setting. Specifically, we set the goal point based on the predicted dynamic objects, and the goal is influenced by the dynamic objects significantly.

In practice, the total potential $U(x)$ is discretized into a grid. Any cell with a value exceeding a chosen threshold (for example, 80–100) is marked as infeasible, thus defining a feasible domain for planning.

To find a viable avoidance point, an RRT-based method explores the feasible region of this potential map starting from the robot's current pose $x_{start}$. Random samples $x_{sample}$ are drawn within map bounds; each sample is rejected if it lands in an infeasible cell or if a direct path from $x_{start}$ to $x_{sample}$ intersects any high-cost region. The samples that pass this feasibility check form a set of candidate points. Although a full tree extension is possible, a simplified variant that directly connects $x_{start}$ to each feasible sample is often sufficient for time-sensitive local planning.

Once a sufficient number of candidates is generated, each candidate $x$ is evaluated using

$$Q(x) = \omega_{safe}S(x) - \omega_{dist}D(x, x_{start}) - \omega_{pot}U(x) \\ - \omega_{hyst}H(x, x_{prev}) \quad (8)$$

Here, $S(x)$ is the safety margin, defined by the distance to the closest obstacle boundary; $D(x, x_{start})$ measures the path cost, typically Euclidean distance; $U(x)$ is the local potential value reflecting both static and dynamic hazards; and $H(x, x_{prev})$ is a hysteresis term penalizing large jumps from the previously chosen avoidance point. The weights $\omega_{safe}$, $\omega_{dist}$, $\omega_{pot}$, $\omega_{hyst}$ should be tuned to balance the priority of safety versus total path cost and planning stability. The point $x_*$ maximizing $Q(x)$ is designated the avoidance point. During a conflict warning, the system immediately directs the robot to $x_*$ to avoid the threat. Upon risk clearance, the robot resumes its original goal or navigation trajectory, thus ensuring active conflict avoidance while minimally impacting the underlying task.

### C. Proactive Conflict Avoidance Mechanism

In human-shared environments, conflicts between human and robotic behaviors can lead to disruptions and safety

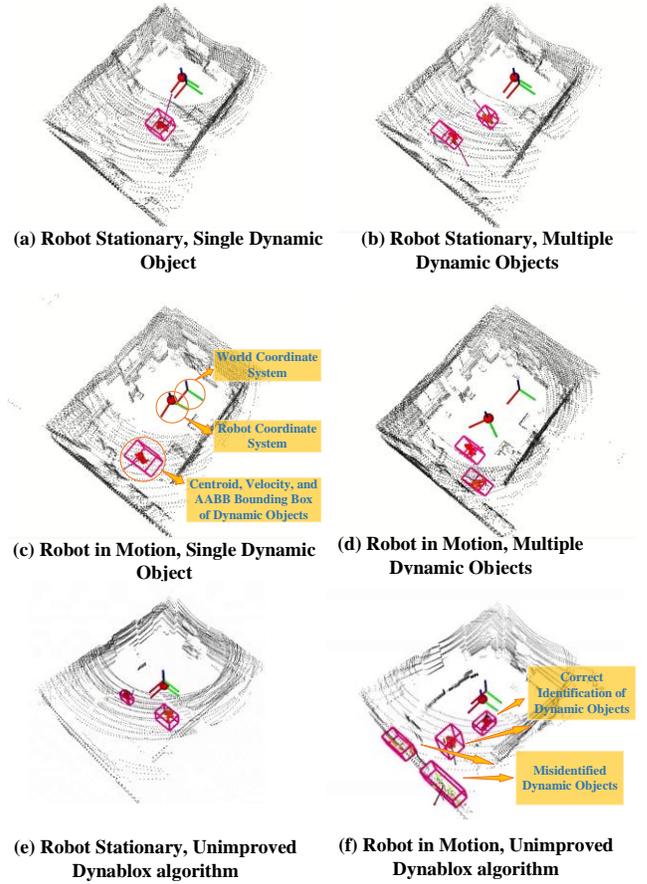

(a) Robot Stationary, Single Dynamic Object
(b) Robot Stationary, Multiple Dynamic Objects
(c) Robot in Motion, Single Dynamic Object
(d) Robot in Motion, Multiple Dynamic Objects
(e) Robot Stationary, Unimproved Dynablox algorithm
(f) Robot in Motion, Unimproved Dynablox algorithm

Fig. 5. Dynamic object detection experiment diagram.

concerns. Relying solely on passive obstacle avoidance is insufficient, as it does not account for proactive conflict resolution when interactions occur. To address this, we propose a proactive behavioral conflict avoidance mechanism for quadruped robots, designed to anticipate and mitigate potential conflicts with humans in real time. The system continuously assesses interaction risks and executes adaptive avoidance strategies before conflicts escalate, ensuring minimal interference with human activities. By integrating this mechanism with an enhanced local planner, the robot maintains smooth task execution through a "perception–adaptation–recovery" framework, allowing it to navigate shared spaces safely and efficiently while preserving operational continuity.

At the core of the proactive conflict avoidance mechanism is the integration of conflict detection with system-level state management. When a conflict risk is detected, the system determines whether the robot is idle or navigating. If idle, it records the current pose for later return; if navigating, it stores the navigation goal to resume once the risk subsides. The avoidance procedure employs local planning methods, such as potential fields or RRT, to generate avoidance points that allow the robot to quickly steer clear of imminent conflicts.

To enhance avoidance efficiency, we improve the TEB method within the local planner. Standard TEB often requires turning maneuvers, which can be time-consuming for quadruped robots. By relaxing the front-back driving constraint, the robot can move forward or backward without

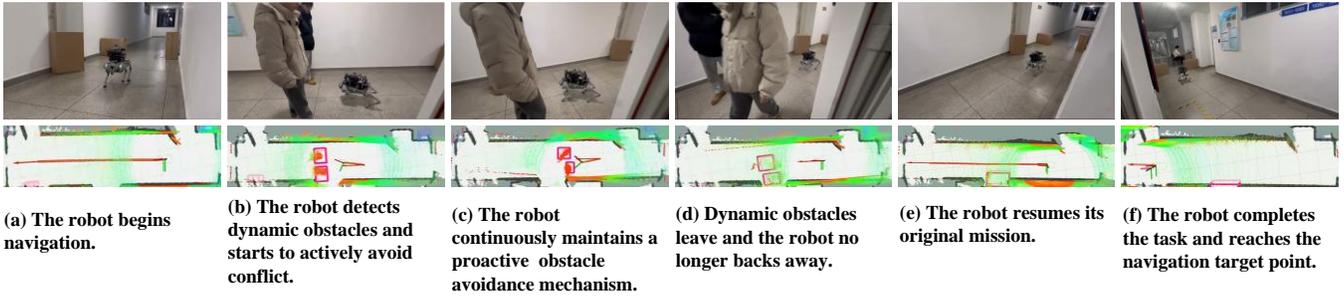

(a) The robot begins navigation.　(b) The robot detects dynamic obstacles and starts to actively avoid conflict.　(c) The robot continuously maintains a proactive obstacle avoidance mechanism.　(d) Dynamic obstacles leave and the robot no longer backs away.　(e) The robot resumes its original mission.　(f) The robot completes the task and reaches the navigation target point.

Fig. 6. Conflict avoidance of pedestrian by robot in mobile working state.

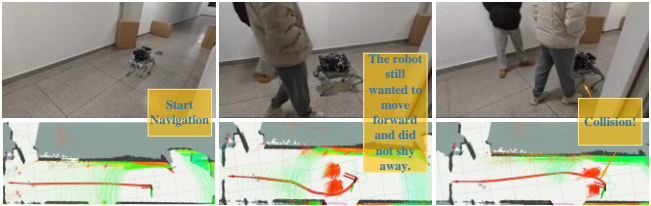

Fig. 7. A conventional method to deal with pedestrian conflict in mobile working state of robot

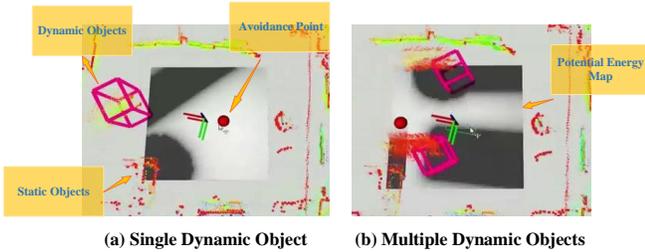

(a) Single Dynamic Object　(b) Multiple Dynamic Objects

Fig. 8. Experimental diagram for the avoidance point selection strategy.
.

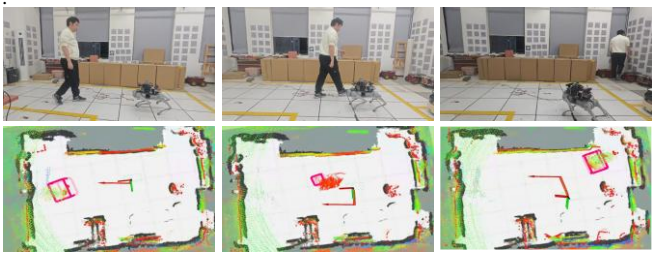

Fig. 9. Conflict avoidance of pedestrian by robot in static working state.

reorienting, enabling faster avoidance of hazardous areas. This modification results in more flexible and compact avoidance paths, particularly beneficial for quadruped robots operating in dynamic or confined environments.

With this framework, the system immediately enters avoidance mode upon detecting a conflict risk, prioritizing safety over the current task. Once the risk is mitigated, it seamlessly resumes its original task.

## V. EXPERIMENTS

The basic carrier of the robot used in this experiment is Unitree GO1 robot, the main control is Intel Core i9-11900H, equipped with a Livox Mid360 LiDAR.

### A. Dynamic Object Detection

To validate the performance of the dynamic object recognition method under different scenarios, this study designed two real-world indoor experimental setups: a stationary robot scenario, in which the robot remains static while one or more dynamic targets are present, and a mobile robot scenario, where the robot moves at a random speed under similar conditions. In the dynamic scenarios, we also compared our approach with the unmodified Dynablox method. As shown in **Figure 5**, the experimental results demonstrate that dynamic targets can be accurately identified regardless of whether the robot is stationary or in motion, and that our method exhibits superior adaptability and interference resistance compared to Dynablox. Moreover, the proposed approach is capable of real-time computation of the motion direction and velocity of dynamic targets, thereby providing reliable data support for subsequent modules.

### B. Avoidance Point Selection Strategy

The experiments in this section serve as validation tests to verify the effectiveness of obstacle prediction and the integration of the potential field map with avoidance point selection. In **Figure 8**, the potential field map and avoidance point selection are illustrated for scenarios in which varying numbers of dynamic objects approach the robot's safety zone while it remains stationary. By accounting for diverse dynamic object situations and the static obstacle environment, the robot can flexibly select reachable avoidance points rather than adhering to a fixed strategy. These results underscore the algorithm's adaptability and robustness.

### C. Proactive Conflict Avoidance Mechanism

This subsection presents a pivotal experiment of this paper, integrating the previously described dynamic object recognition and avoidance point selection strategies. By employing an improved TEB for local path planning, our method enables the robot to perform active safety conflict avoidance. To validate the effectiveness of our approach, two scenarios were designed. The first scenario involves a stationary robot in an indoor space that must promptly yield to dynamic obstacles and subsequently resume its task. The second scenario examines a robot navigating a corridor while similarly executing timely conflict avoidance and task recovery. For the corridor scenario, comparative experiments were conducted with and without the proactive conflict avoidance L.

**Figure 9** illustrates the proactive safety conflict avoidance performance when the robot is stationary. The robot can anticipate the movement of dynamic objects and, by considering its surroundings, quickly select an appropriate avoidance point. Once the dynamic obstacle leaves the predefined safe zone, the robot promptly returns to its original position. To the best of our knowledge, this mechanism, where a stationary mobile robot without an active navigation task or designated target can proactively avoid obstacles and resume its tasks, is rarely explored. Unlike conventional dynamic obstacle avoidance techniques in unmanned aerial vehicles that rely on predetermined trajectories, our approach dynamically incorporates static environmental factors, highlighting its advantages.

**Figures 6** and **Figures 7** illustrate the robot's navigation performance in a corridor environment. As shown in **Figure 7** without the proactive conflict avoidance mechanism, the robot remains focused on reaching its target. In such cases, if a human does not yield or if the planned path becomes impassable, the robot may enter into conflict with the human, creating potential safety risks. In contrast, with the proposed algorithm, the robot proactively retreats until the dynamic obstacle has cleared before seamlessly resuming its navigation task. **Figure 6** further demonstrates the robot's ability to anticipate the movement of dynamic obstacles, enabling it to execute preemptive safety maneuvers and avoid potential collisions. By integrating the proactive conflict avoidance mechanism, the robot not only enhances operational safety in dynamic environments but also ensures smoother task execution while minimizing disruption to human activities.

## VI. Conclusion

In this paper, we presented a proactive conflict avoidance framework for quadruped robots operating in shared human–robot environments. By integrating robust dynamic object detection, a novel avoidance point selection strategy, and an enhanced local planner, our approach enables the robot to not only maintain its own safety but also minimize interference with human activities. Experimental results in both stationary and mobile scenarios validate the robustness of the framework: in a stationary setting, the robot effectively detects oncoming dynamic obstacles, proactively retreats, and returns to its initial pose when safe; in motion, the robot is able to disengage from its navigation goal, yield to pedestrians in confined corridors, and rejoin its planned route when hazards subside. Overall, the proposed conflict avoidance system advances the goal of seamless coexistence between humans and quadruped robots by achieving reliable conflict avoidance while maintaining high task efficiency.